\documentclass{article}
\newcommand{\methodname}{Video-Infinity}

\newcommand{\attnname}{Dual-scope attention}

\newcommand{\pname}{Clip parallelism}

\newcommand{\codeform}[1]{\texttt{#1}}

\PassOptionsToPackage{numbers, compress}{natbib}

\usepackage[preprint]{src/neurips_2024}

\usepackage[utf8]{inputenc} 
\usepackage[T1]{fontenc}    
\usepackage{hyperref}       
\usepackage{url}            
\usepackage{booktabs}       
\usepackage{amsfonts}       
\usepackage{nicefrac}       
\usepackage{microtype}      
\usepackage{xcolor}         
\usepackage{svg}
\usepackage{amsmath}
\usepackage{algorithm}
\usepackage{algorithmic}
\usepackage{enumitem} 
\usepackage{wrapfig}
\usepackage{graphicx}
\usepackage{wrapfig}

\title{Video-Infinity: Distributed Long Video Generation}

\author{
    \vspace{-2em}
    \\
    \href{https://video-infinity.tanzhenxiong.com/}{
    https://video-infinity.tanzhenxiong.com
    }
    \\
    Zhenxiong Tan
    \footnotemark[1]
    \quad 
    Xingyi Yang
    \thanks{Equal Contribution}
    \quad 
    Songhua Liu
    \quad
    Xinchao Wang
    \thanks{Corresponding Author} 
    \\
    National University of Singapore \\
    \texttt{zhenxiong@u.nus.edu \quad xinchao@nus.edu.sg} \\
}

\begin{document}

{
\maketitle
}

\vspace{-1em}
\begin{figure}[h]
    \centering
    \includegraphics[width=0.95\textwidth]{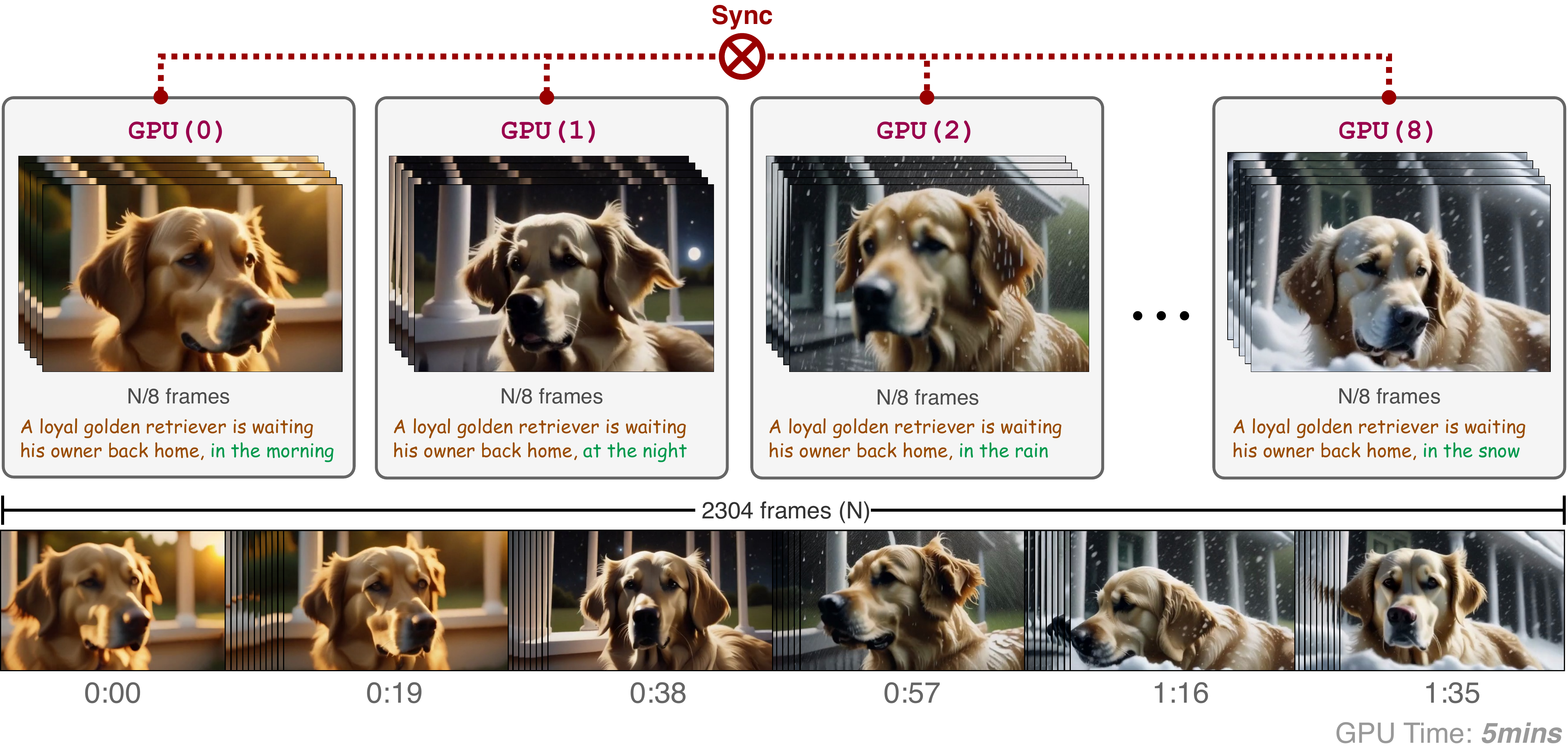}
    \vspace{-1em}
    \caption{
        Multiple GPUs parallelly generate a complete video, producing 2300 frames in 5 minutes.
    }
    \label{fig:example}
\end{figure}

\begin{abstract}
    

Diffusion models have recently 
achieved remarkable results
for video generation.
Despite the encouraging performances, 
the generated videos are typically
constrained to a small number of frames, 
resulting in clips lasting merely a few seconds. 
The primary challenges in producing longer videos 
include the substantial memory requirements
and the extended processing time required on a single GPU.
A straightforward solution would be to 
split the workload across multiple GPUs,
which, however, leads to two issues:
(1) ensuring all GPUs communicate 
effectively to share timing and context information,  
and (2) modifying existing video diffusion models, 
which are usually trained on short sequences, 
to create longer videos without additional training. 
To tackle these, in this paper we introduce {\methodname},
a distributed inference pipeline that 
enables parallel processing across multiple GPUs 
for long-form video generation. 
Specifically, 
we propose two coherent mechanisms: \emph{\pname} and \emph{\attnname}. {\pname} optimizes the gathering and sharing of context information across GPUs 
which minimizes communication overhead,
while {\attnname} modulates the temporal self-attention to balance local and global contexts efficiently across the devices. 
Together, the two mechanisms join forces
to distribute the workload 
and enable the fast generation of long videos.
Under an 8 $\times$ Nvidia 6000 Ada GPU (48G) setup, our method generates videos up to 2,300 frames in approximately 5 minutes,
enabling long video generation at a speed 100 times faster than the prior methods.



\end{abstract}


\section{Introduction}
A long-standing pursuit of human being is to replicate the dynamic world we live in, in the digital system. 
Traditionally dominated by physics and graphics, this effort has recently been enhanced by the emergence of data-driven generative models~\cite{rombach2022high,ho2022video, harvey2022flexible, ho2022imagen}, which can create highly realistic images and videos indistinguishable from reality. 
However, these models typically produce very short video segments, with most limited to 16-24 frames~\cite{guo2023animatediff, chen2023videocrafter1, chen2024videocrafter2}. 
Some models extend to 60 or 120 frames~\cite{arc2024mira, hpcaitech2024opensora}, but compromise heavily on resolution and visual quality.  


Generating long video poses substantial challenges, primarily due to the extensive resource demands for model training and inference. 
Current models, constrained by available resources, are often trained on brief clips, making it difficult to sustain quality over longer sequences.
Moreover, generating a minute-long video
in one go can overwhelm GPU memory, making the task seem elusive.

Existing solutions, including autoregressive, hierarchical, and short-to-long methods, offer partial remedies but have significant limitations.  
Autoregressive methods \cite{henschel2024streamingt2v, yin2023nuwa} produce frames sequentially, dependent on preceding ones.
Hierarchical methods ~\cite{chen2023seine, yin2023nuwa, zhou2024storydiffusion} create keyframes first, then fill in transitional frames.
Furthermore, some approaches treat a long video as multiple overlapping short video clips~\cite{qiu2023freenoise, wang2023gen}. 
These methods are not end-to-end; they often miss global continuity, require extensive computation, especially in regions of overlap, and struggle with consistency across segments.


To bridge these gaps, we introduce a novel framework for distributed long video generation, termed \methodname. 
On the high level, it work in a divide-and-conquer principle. It breaks down the task of long video generation into smaller, manageable segments. 
These segments are distributed across multiple GPUs, allowing for parallel processing. All clients should work collaboratively to ensure the final video is coherent in semantics.

This setup, while straightforward, faces two principal challenges: ensuring effective communication among all GPUs to share 
and contextual information, and adapting existing models—typically trained on shorter sequences—to generate longer videos without requiring additional training.


To overcome these challenges, we introduce two synergistic mechanisms: \emph{\pname} and \emph{\attnname}.
\pname~enables efficient collaboration among multiple GPUs 
by splitting contextual information into three parts. 
It uses an interleaved communication strategy 
to complete the sharing in three steps. 
Building on the capabilities of \pname, \attnname meticulously adjusts the temporal self-attention mechanisms to achieve an optimal balance between local and global contexts across devices. This balance allows a model trained on short clips to be extended to long video generation with overall coherence.


Even more exciting, by leveraging both strategies, Video-Infinity reduces memory overhead from a quadratic to a linear scale. With the power of multiple device parallelism and sufficient VRAM, our system can generate videos of any, potentially even infinite length.

As a results, our method significantly extends the maximum length of videos that can be generated and accelerates the speed of long video generation.
Specifficly, on an 8 $\times$ Nvidia 6000 Ada (48G) setup,  our method manages to generate videos up to 2300 frames in just 5 minutes.
Our contributions are summarized as follows: 
(1) We are the first to address long video generation using distributed parallel computation, enhancing scalability and reducing generation times. 
(2) We introduce two interconnected mechanisms: \pname, which optimizes context information sharing across GPUs, and \attnname, which adjusts temporal self-attention to ensure video coherence across devices.
(3) Our experiments show that, compared to the existing ultra-long text-to-video method Streaming T2V~\cite{henschel2024streamingt2v}, our approach can be up to 100 times.


\section{Related works}
\subsection{Diffusion models}
Diffusion models have gained significant attention in recent years due to their impressive ability to generate high-quality media. 
Originally introduced for image synthesis, models like Denoising Diffusion Probabilistic Models (DDPM)~\cite{ho2020denoising} and Latent Diffusion Models (LDM)~\cite{rombach2022high} have demonstrated state-of-the-art performance in image generation. 
These models progressively denoise a Gaussian noise distribution by learning a sequence of reverse transformations. 
Beyond images~\cite{ho2020denoising,rombach2022high}, diffusion models have also shown promise in audio~\cite{kong2020diffwave, yang2023diffsound,liu2023audioldm} and 3D generation~\cite{luo2021diffusion, poole2022dreamfusion}.
Adaptations of diffusion models for video generation incorporate temporal modules to capture the sequential nature of video frames. 
For instance, Video Diffusion Models (VDM)~\cite{ho2022video} and Flexible Diffusion Model (FDM)~\cite{harvey2022flexible} effectively extend diffusion frameworks to video data, overcoming challenges like temporal consistency and quality degradation. 
More recent models such as AnimateDiff~\cite{guo2023animatediff}, ModelScope~\cite{wang2023modelscope}, and VideoCrafter~\cite{chen2023videocrafter1, chen2024videocrafter2} can now produce video clips with better dynamics and improved visual quality.

\subsection{Techniques for long video generation}
Streaming T2V~\cite{henschel2024streamingt2v} introduces a method that relies on a conditional attention module to ensure smooth transitions between video segments and a scene-preserving mechanism for content consistency. 
However, this method requires training and is not end-to-end, posing limitations on its practicality.
FreeNoise~\cite{qiu2023freenoise} utilizes rescheduled noise sequences and window-based temporal attention to improve video continuity. Despite these innovations, the rescheduled noise contributes to limited dynamics in the generated videos, and the overlapping attention windows introduce additional computational overhead.
NUWA-XL~\cite{yin2023nuwa} from the NUWA series employs a ``coarse-to-fine'' autoregressive approach, using a global diffusion model to generate keyframes and local models to fill the intermediate frames. 
Although promising, NUWA-XL has been trained only within a narrow domain and has not yet made its models and code available, limiting both its evaluation and reproducibility.
Gen-L-Video~\cite{wang2023gen} adapts short video diffusion models to handle long videos conditioned on multiple texts without requiring additional training. 
This approach cleverly uses latent overlaps to extend video length, which is a common strategy among recent methodologies.
SEINE~\cite{chen2023seine} leverages a random-mask diffusion method to automate the generation of transition videos between scenes, guided by textual descriptions. 
Like other models, SEINE employs an autoregressive approach and requires image conditioning to facilitate the generation process.


\subsection{Distributed diffusion}
Recently, to reduce the latency of each denoising step in diffusion models, various distributed parallel methods have been applied to image diffusion models.
ParaDiGMS~\cite{shih2024parallel} utilizes step-based parallelism, where each denoising step is executed on a different GPU device in parallel.
However, this approach tends to waste much computation. 
Another method, DistriFusion~\cite{li2024distrifusion}, employs a technique of dividing images into patches, allowing different patches to be denoised on separate GPUs. 
This approach ensures synchronization among patches and achieves minimal computational waste. 
However, it is designed specifically for image diffusion and requires significant communication overhead and specialized hardware support to achieve low latency.


\section{Preliminaries}


\textbf{Diffusion Models in Video Generation}

The process of generating videos using diffusion models involves progressively denoising the latent representation, denoted as $x_t$, where $t$ ranges from 0 to $T$. 
The initial noisy video latent is represented by a random noise tensor $x_T$. 
With each denoising step, $x_t$ is updated to a clearer latent $x_{t-1}$. 
This iterative process continues until $x_T$ is denoised to $x_0$, which is then fed into a decoder to generate the final video.
The key aspect of updating $x_t$ to $x_{t-1}$ is computing the noisy prediction $\epsilon_t$, given by:
\begin{equation}
\epsilon_t = \mathcal{E}_\theta(x_t),
\end{equation}
where $\mathcal{E}_\theta$ represents the diffusion model. 

The diffusion model $\mathcal{E}_\theta$ can be implemented using various architectures, such as U-Net~\cite{ronneberger2015u,ho2022video, harvey2022flexible, guo2023animatediff,chen2023videocrafter1} or DiT~\cite{peebles2023scalable, hpcaitech2024opensora, arc2024mira}.
These diffusion models are generally composed of several similar layers. 
More specifically, the initial random noise tensor is written as $x_T \in \mathbb{R}^{F \times H \times W \times C}$, where $F$ represents the number of frames, $H$ and $W$ denote the height and width of each frame, respectively, and $C$ is the number of channels.

The latent tensor $v$ in each layer generally maintains a consistent shape, $v \in \mathbb{R}^{F \times H' \times W' \times C'}$, where $F$  remains constant across layers. 
The dimensions $H'$, $W'$, and $C'$ can vary due to the down-sampling and up-sampling operations of the U-Net architecture.

These layers in the diffusion model $\mathcal{E}_\theta$  are usually composed of two main types of modules: spatial and temporal.
The spatial modules receive slices of the latent $v$ shaped $v \in \mathbb{R}^{H' \times W' \times C'}$ (a single frame), representing tokens for each video frame in the latent space.
They independently process spatial features within each frame.
The temporal modules receive elongated strips of the latent tensor $v$ shaped $v \in \mathbb{R}^{F \times C'}$, representing tokens containing temporal information across frames at specific spatial locations. 
They capture temporal dependencies between frames at each location.


\section{Distributed Long Video Generation}

\begin{figure}[t]
    \centering
    \includegraphics[page=1, width=\textwidth]{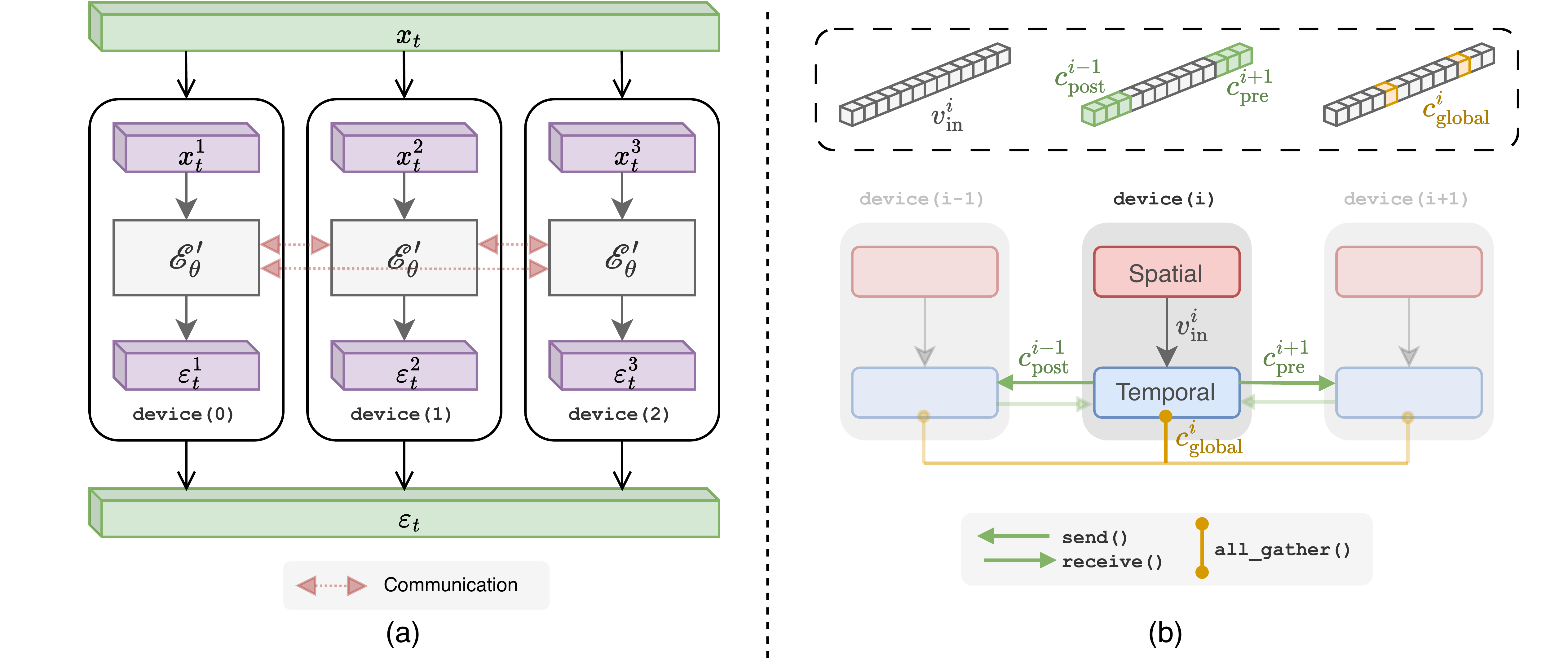}
    \caption{
     (a) \textbf{Pipeline of \methodname}: 
     The latent tensor is split into clips and distributed to different devices. 
     The diffusion model predicts noise in parallel with communication, and the noises are concatenated to produce the final output.
     (b) \textbf{Illustration of \pname}: In each layer of the video diffusion module, spatial modules operate independently, whereas temporal modules synchronize context elements \(c^i_{\text{pre}}\), \(c^i_{\text{post}}\), and \(c^i_{\text{global}}\). Peer-to-peer and collaborative communications are employed.}
    \label{fig:communication}
    \vspace{-1em}
\end{figure}




At the core of our pipeline, \methodname~segments the video latent into chunks, which are then distributed across multiple devices.
An overview of our method is shown in Figure~\ref{fig:communication}, where we divide the video latent along the temporal dimension. Such partitioning allows for parallel denoising on different devices, each handling non-overlapping frames. 
To facilitate this, we propose \pname, detailed in  in Section~\ref{sec:clipp}, a mechanism that efficiently synchronizes temporal information across devices. Additionally, we incorporate \attnname~in Section~\ref{section:module_parallel}, which modulates temporal attention to ensure training-free long video coherence.

Formally, \methodname~splits the noisy latent $x_T \in \mathbb{R}^{F \times H \times W \times C}$ into $N$ sub-latent clips $x_T^i \in \mathbb{R}^{F_\text{clip} \times H \times W \times C}$, where $i \in [1, N]$, $F_\text{clip} = {F}/{N}$ represents the number of frames in each clip, and $N$ represents the total number of clips. This structured segmentation facilitates an even load distribution across $N$ devices. Additionally, the spatial modules of video diffusion models operate independently across frames, which eliminates the need for inter-device communication and maintains consistency in the outputs across different devices.


\subsection{\pname~for video diffusion}
\label{sec:clipp}

To ensure coherence among clips distributed on different devices, we propose \pname,  shown in Figure~\ref{fig:communication}. It parallelizes the temporal layers for video diffusion models and enables efficient inter-device communication.


\textbf{Parallelized temporal modules.}
In the standard diffusion model, a temporal module aggregates features across frames, which could be simplified as 
\begin{equation}
    v_{\text{out}} = \text{temporal}\left(v_{\text{in}}\right),
\end{equation}
where $v_{\text{in}}\in \mathbb{R}^{F\times * \times C'}$ is the input feature of this temporal layer.

However, \methodname~distributes input feature tensors $v_{\text{in}}$ across multiple devices, dividing them into several clips $v_{\text{in}}^{i} \in \mathbb{R}^{{F}_\text{clip} \times *\times C'}$, each placed on \codeform{device(i)}. To facilitate distributed inference \emph{without modifying the original structure} of the temporal modules, we redefine the temporal operation. This modified operation now considers not only the current clip, but also adjacent clips and global semantics. Conceptually, the parallelized temporal modules are defined as follows:
\begin{align}
    v_{\text{out}}^{i} &= 
    \text{temporal}_{\text{Parallel}}\left(
        v_{\text{in}}^{i},
        c^i
    \right),\\
    c^i &= \left\{
     c^{i}_{\text{pre}} ,
     c^{i}_{\text{post}} ,
     c_{\text{global}} 
    \right\}
\end{align}
where \( c^{i} \) stands for the temporal information that enriches each device's computation by incorporating inter-device context. 
Each \( c^{i} \) includes temporal information from the preceding \codeform{device(i-1)} via \( c^{i}_{\text{pre}} \), and from the succeeding \codeform{device(i+1)} via \( c^{i}_{\text{post}} \).
Furthermore, \( c_{\text{global}} \) is a selective aggregate of inputs from all devices, optimizing global information coherence and reducing overhead.

The output for each device, $v_{\text{out}}^{i}$, reflects localized computations augmented by these contextual inputs.
The complete output of the layer, $v_{\text{out}}$, is obtained by concatenating the outputs from all devices:
\begin{equation}
 v_{\text{out}} = \text{Concat} \left( \left\{v_{\text{out}}^{i}|i \in \left[1, N\right]\right\}\right)
\end{equation}
This concatenation provides a holistic view of the processed features, maintaining temporal coherence across the distributed system. Further details on how these temporal modules integrate context will be discussed in Section~\ref{section:module_parallel}.






\begin{wrapfigure}{r}{0.42\textwidth}
\vspace{-3mm}
    \centering
    \includegraphics[page=1, width=0.46\textwidth]{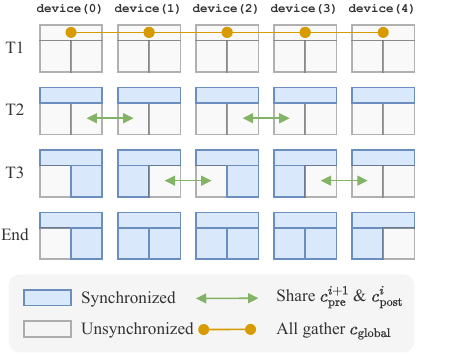}
    \vspace{-2mm}
    \caption{Tree different stages in the communication process of \pname}
    \label{fig:communication}
    \vspace{-8mm}
\end{wrapfigure}
\textbf{Three-round context communication.} Redefining the temporal modules necessitates efficient communication of the context components $c^i = \left\{
     c^{i}_{\text{pre}} ,
     c^{i}_{\text{post}} ,
     c_{\text{global}} 
    \right\}$. This is achieved through a three-stage synchronization process, where each stage addresses a specific part of the context, as illustrated in Figure~\ref{fig:communication}.

In the first stage, \textbf{T1}, each \codeform{device(i)} broadcasts its global context \(c^i_{\text{global}}\) with all other devices through an \codeform{all\_gather()} operation. This operation ensures that every device receives the global context, maintaining global consistency across the entire video.

The subsequent stages, \textbf{T2 and T3} focus on exchanging neighboring contexts. Due to connection limits\footnote{Only one device can communicate with another at a time.}, we employ an interleaved strategy. In T2, odd-numbered nodes send their \(c^{i+1}_{\text{pre}}\) to their subsequent  \codeform{device(i+1)}, and even-numbered nodes send their \(c^{i-1}_{\text{post}}\) to \codeform{device(i-1)}. 
In T3, this pattern reverses—odd-numbered devices receive context from their predecessors, and even-numbered devices from their successors. This approach prevents bottlenecks, optimizes channel usage, and minimizes deadlock risks.

Finally, all nodes complete context synchronization, ensuring that each device has the full context needed to perform its computations. More details can be found in the pseudocode in Appendix~\ref{ap:algorithm}.

\subsection{Putting each module in parallel}
\label{section:module_parallel}
Building upon the \pname, this section details that how information is synchronized in each temporal module. A key technique here is the \emph{\attnname}, which facilitates training-free long video generation and reduces the communication cost.

There are typically three temporal modules in video diffusion models: attention module~\cite{vaswani2017attention} \codeform{Attention()}, convolution module~\cite{o2015introduction} \codeform{Conv()}, and group normalization module~\cite{wu2018group} \codeform{GroupNorm()}. We have tailored these modules to integrate into \pname, enabling distributed processing across multiple devices for efficient and coherent video content synchronization.


\textbf{DualScope attention.}
Applying attention in parallel inference incurs new challenges. 
The original attention module require simultaneous access to all input tokens~\cite{shaw2018self}.
To adopt it under \pname, it necessitates aggregating tokens across devices,  resulting in tremendous  communication costs.
Additionally, those attention trained on shorter video clips often degrade in quality when applied to longer sequences.

To address these issues, we introduce the \emph{DualScope attention} module. It revises the computation of K-V pairs to incorporate both local and global contexts into the attention. For each query token from frame $a$, its corresponding keys and values are computed from tokens in the frame set $\mathcal{A}^a = \mathcal{N}^a \cup \mathcal{G}$:
\begin{itemize}
    \item \textit{Local Context ($\mathcal{N}^a$).} This includes the $|\mathcal{N}^a|$ neighboring frames of $a$, from which the keys and values are derived to capture the local context. This local setup is typically achieved through window attention, focusing on the nearby frames to enhance the temporal coherence.
    \item \textit{Global Context ($\mathcal{G}$).} In contrast, the global context consists of frames uniformly sampled from videos across all devices. This context provides keys and values from a broader range, giving the model access to long-range information
\end{itemize}
In practice, the keys $K$ and values $V$ are constructed by concatenating the tokens from both contexts $K=\text{Concat}(K_\text{local},K_\text{global})$ and $V=\text{Concat}(V_\text{local},V_\text{global})$, where $K_\text{local}$ and $Q_\text{local}$ is derived from $\mathcal{N}^a$ and $K_\text{global}$ and $Q_\text{global}$ from $\mathcal{G}$. We find that this modified key-value computation can be easily incorporated into existing temporal attention without additional training, enhancing the coherence of long videos.

In the implementation of \pname, above reformulated attention largely reduce the communication overhead. Comparing to gathering all tokens of length $F$, we only synchronize constant number of tokens. Specifically, we set  $|c^{i}_{\text{pre}}| = |c^{i}_{\text{post}}|=\frac{|\mathcal{N}^a|}{2}$ and $|c_{\text{global}}| = |\mathcal{G}|$, with both $|\mathcal{N}^a|$ and $|\mathcal{G}|$ configured to 16. This reduces data synchronization demands while still capturing essential local and global information.



\textbf{Convolution module.}
The temporal convolution module \codeform{Conv()} applies convolution along the temporal dimension to its input $v_{\text{in}}^i \in \mathbb{R}^{F_\text{clip} \times C'}$. 
In \pname, the context $c^i$ of the \codeform{Conv()} includes \( c^{i}_{\text{pre}} \) and \( c^{i}_{\text{post}} \). They are padded to the original sequences. 
Specifically, \( c^{i}_{\text{pre}} \) consists of the last \( n \) frames of \( v_{\text{in}}^{i-1} \), and \( c^{i}_{\text{post}} \) consists of the first \( n \) frames of \( v_{\text{in}}^{i+1} \), where \( n \) is the receptive field size of the convolution.

\textbf{Group normalization.} 
In video diffusion model, group normalization is applied to the input tensor $v_{\text{in}}^i \in \mathbb{R}^{F_\text{clip} \times H \times W \times C'}$ to maintain consistent feature scaling across different frames. 

In \pname, each device first computes the group mean $\mu^i$ of its respective video clip. These means are aggregated to compute the global mean  $\bar{\mu}=\frac{\sum_{i=1}^N \mu^i}{N}$, where N is the number of devices. 
Subsequently, using $\bar{\mu}$, each device computes its standard deviation $\bar{\sigma}^i$, which is shared to calculate the global standard deviation $\bar{\sigma}$. 
The global mean $\bar{\mu}$ and global standard deviation $\bar{\sigma}$, serving as $c_{\text{global}}$, are used for normalization~\footnote{Note that simply averaging the individual standard deviations $\sigma^i$ does not yield the true global standard deviation $\bar{\sigma}$.}. 

\section{Experiments}

\subsection{Setups}
\textbf{Base model.}
In the experiments, the text to video model VideoCrafter2~\cite{chen2024videocrafter2} (320 x 512) is selected as the base model of our method.
VideoCrafter2, which was trained on 16-frame videos, excels at generating video clips that are both consistent and of high quality. It is also the highest-scoring open-source video generation model under the VBench~\cite{huang2023vbench} evaluation, achieving the top total score.

\textbf{Metrics evaluation.}
VBench~\cite{huang2023vbench} is utilized as a comprehensive video evaluation tool, featuring a broad array of metrics across various video dimensions. For each method, videos are generated using the prompts provided by VBench for evaluation. 
The metrics measured encompass all the indicators under the Video Quality category in VBench, including subject consistency, background consistency, temporal flickering, motion smoothness, dynamic degree, aesthetic quality and imaging quality.
Given that VBench's evaluation is typically performed on video clips of 16 frames, we have modified the evaluation method for videos longer than 16 frames: we randomly sample five 16-frame clips from each video to evaluate separately, and then calculate the average score of these assessments.

\textbf{Baslines.}
Our approach is benchmarked against several other methods:
\begin{itemize}[leftmargin=0.5cm]

\item \textbf{FreeNoise~\cite{qiu2023freenoise}}:
We chose FreeNoise as a baseline because it is also a training-free method that can base the VideoCrafter2~\cite{chen2024videocrafter2} model, which also serves as our base model, to generate long videos. 
It employs a rescheduling technique for the initialization noise and incorporates Window-based Attention Fusion to generate longer videos.

\item \textbf{Streaming T2V~\cite{henschel2024streamingt2v}}:
To assess our method's effectiveness in generating longer videos, StreamingT2V was chosen as our baseline. 
Streaming T2V involves training a new model that uses an auto-regressive approach to produce long-form videos.
Like our approach, it also has the capability to generate videos exceeding 1000 frames.

\textbf{OpenSora V1.1}~\cite{hpcaitech2024opensora}, a video diffusion model based on DiT~\cite{peebles2023scalable}, supports up to 120 frames, can generate videos at various resolutions, and has been specifically trained on longer video sequences to enhance its extended video generation capabilities.

\end{itemize}

\textbf{\attnname setting.}
In the implementation of the \attnname, the number of neighboring frames $\mathcal{N}^i$ is set to 16, with 8 frames coming from the preceding clip and 8 frames from the subsequent clip. 
The number of global frames, \( \mathcal{G} \), is set to 16.  
To balance consistency and dynamics during the denoising process, the weights of frames in \( \mathcal{G} \) and \( \mathcal{N}^i \) are dynamically adjusted. 
Specifically, the weight of \( \mathcal{G} \) increases by 10 for timesteps  $t$ greater than 800, whereas the weight of \( \mathcal{N}^i \) increases by 10 for timesteps $t$ less than or equal to 800.


\textbf{Implementation details.}
By default, all parameters of the diffusion are kept consistent with the original inference settings of VideoCrafter2~\cite{chen2024videocrafter2}, with the number of denoising steps set to 30. 
Our experiments are conducted on 8 \( \times  \) Nvidia 6000 Ada (with 48G memory) . 
To implement the temporal module in \pname, we utilized the \codeform{torch.distributed} tool package, employing Nvidia's NCCL as the backend to facilitate efficient inter-GPU communication.
Additionally, all fps conditions are set to 24, and the resolution is set to $512 \times 320$. Note that the resolution for Streaming T2V cannot be modified; thus, videos are generated at its default resolution ($256 \times 256$ for preview videos and $720 \times 720$ for final videos).

\begin{wraptable}{h}{0.5\textwidth}
\resizebox{\linewidth}{!}{
    \centering
    \small
        \begin{tabular}{l|c|cc}
        \toprule
        \textbf{Method} & \textbf{Max} & \multicolumn{2}{c}{\textbf{Time Coast (second)}}\\
                        & \textbf{Frames} & \textbf{128 frames}&\textbf{1024 frames}\\ 
        \midrule
        Ours            &                          \textbf{2300}&                        \textbf{21}&\textbf{131}\\
        ST2V (preview)&                          -&                        277&2,196\\
     ST2V (final cut)& -& 1730&13,726\\
        FreeNoise~\cite{qiu2023freenoise}&                          128&                        201&$\times$\\
        Open-Sora v1.1~\cite{hpcaitech2024opensora}       & 120&                        234&$\times$\\ 
        \bottomrule
        \end{tabular}
        }
        \caption{Comparison of maximum frames and generation times for different methods.}
\label{tab:capacity}
\vspace{-1em}
\end{wraptable}

\subsection{Main results}

\textbf{Capacity and efficiency.}

We evaluated the capabilities of our method on an 8 $\times$ Nvidia 6000 Ada (48G) setup. 
Our approach successfully generated videos of \textbf{2300 frames} at a resolution of 512 $\times$ 320, equivalent to a duration of 95 seconds at 24 frames per second. 
Remarkably, the entire computation process took approximately \textbf{5 minutes} (312s), benefiting from efficient communication and the leveraging of multi-GPU parallel processing. 

Table~\ref{tab:capacity} presents the capacities for long video generation of various methods, all measured under the same device specifications. 
To ensure comparability, we standardized the resolution of the videos generated by all methods to 512x320. 
For StreamingT2V, we provide two sets of data: one for generating preview videos at 256x256 resolution, and another for final videos produced at a resolution of 720x720.
The results demonstrate that our method is the most capable within the end-to-end category, generating the longest videos of up to 2300 frames — 8.2 times more than OpenSora V1.1. 
Additionally, our method consistently produces the final videos in the shortest time, both for short videos of 128 frames and long videos of 1024 frames.
Notably, in the generation of 1024-frame videos, our method is over 100 times faster than StreamingT2V, the only baseline method capable of producing videos of this length. 
Even when compared to the speed of generating smaller, lower-resolution preview videos by StreamingT2V, our method is 16 times faster.

\begin{figure}[t]
    \centering
    \includegraphics[page=1, width=0.95\textwidth]{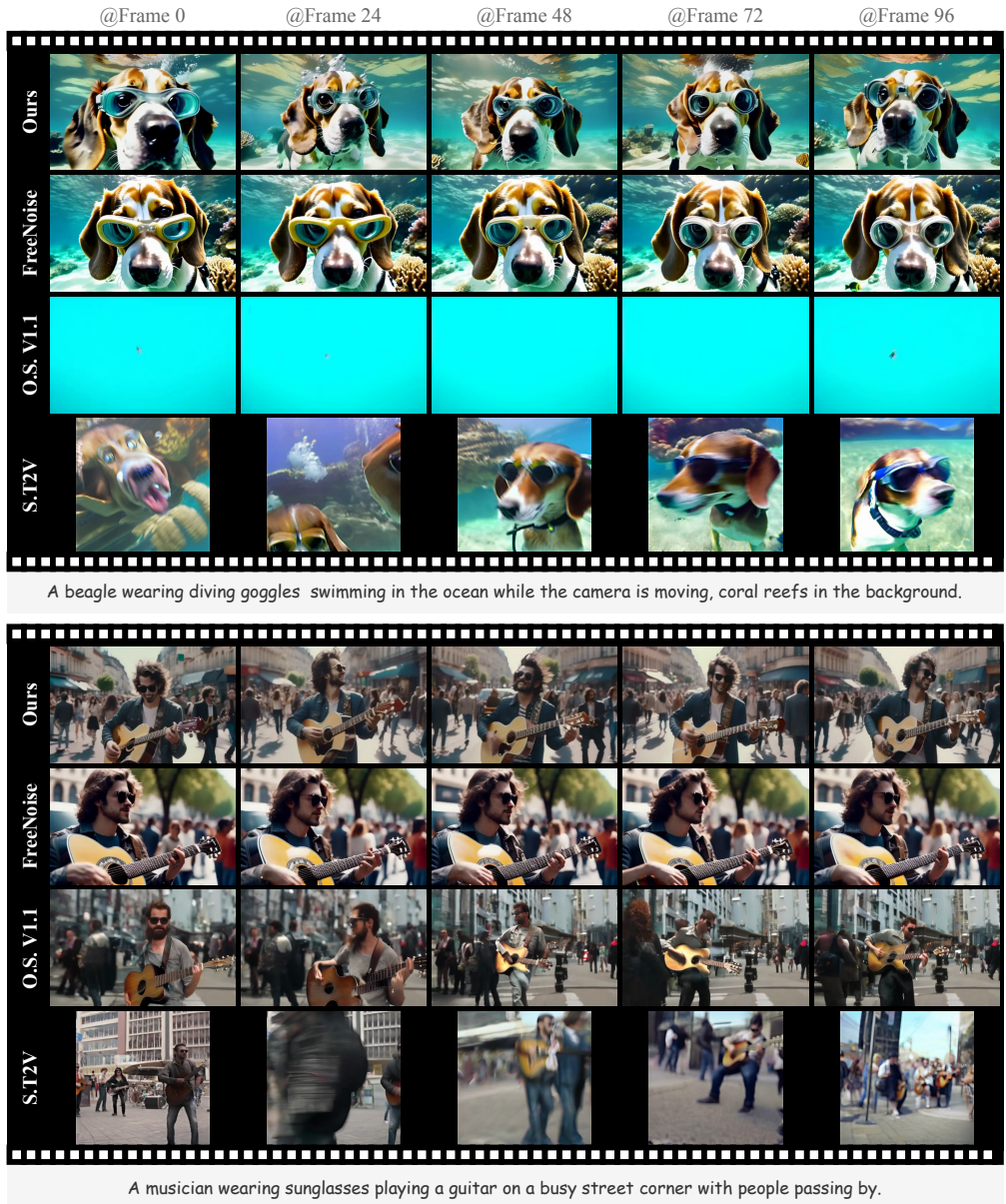}
    \caption{
   Comparison of frame images from sample videos generated by different methods. 
    }
    \label{fig:comparision}
     \vspace{-1em}
\end{figure}

\textbf{Video quality.}
We compared the videos generated by our method with those produced by FreeNoise~\cite{qiu2023freenoise}  and StreamingT2V~\cite{henschel2024streamingt2v} for long video generation. 
Figure~\ref{fig:comparision} visualizes some frames from videos generated by different methods using the same prompt. 
Additionally, Table~\ref{tab:metrics} displays the quality of the videos produced by these methods, evaluated across various metrics in VBench~\cite{huang2023vbench}.

Figure~\ref{fig:comparision} shows that while the StreamingT2V~\cite{henschel2024streamingt2v} method generates long videos with sufficient dynamism, they lack consistency between the beginning and end.
Conversely, videos generated by FreeNoise~\cite{qiu2023freenoise} maintain consistency in object placement throughout but exhibit minimal variation in visuals. 
For example, as shown in Figure~\ref{fig:comparision}, the video of the person playing the guitar maintains a single pose with only minimal movement. 
Similarly, the dog on the left remains intently focused on the camera, with no changes in the position of its ears, nose, or body.
OpenSora V1.1~\cite{hpcaitech2024opensora} failed to generate the first video and the second video's background was not smooth.
In contrast, our method not only ensures better consistency but also features more significant motion in the generated videos.

\begin{table}[t]
    \resizebox{\linewidth}{!}{
        \centering
        \begin{tabular}{l|c|ccccccc|c}
        \toprule
             \textbf{Method} &\textbf{Number of}&  \textbf{Subject}     &  \textbf{Background}   & \textbf{Temporal}      & \textbf{Motion}        & \textbf{Dynamic}   & \textbf{Aesthetic} & \textbf{Imaging}  &\textbf{Overall}\\
              &\textbf{Frames}&  \textbf{consistency}&  \textbf{consistency}  & \textbf{flickering}    & \textbf{smoothness}    & \textbf{degree}    & \textbf{quality}   &\textbf{quality}  &\textbf{Score}\\
         \midrule
         \midrule
             
 \textbf{V.C.2}&16& 96.85\%& 98.22\%&98.41\% &97.73\% & 42.50\%& 63.13\% & 67.22\% &80.58\%\\
         \midrule
        \textbf{FreeNoise}    &64&  94.16\%& \textbf{96.63\%}& 98.37\% & 97.04\% & 44.44\% &  \textbf{60.53\%}& 67.44\%  &79.80\%\\
        \textbf{OpenSora v1.1}    &64&  86.18\%& 95.83\%& \textbf{98.47\%}& \textbf{97.27\%}& 73.61\%&  51.69\%& 50.61\%&79.09\%\\
        \textbf{Ours}    &64&  \textbf{97.77\%}&  93.90\%&  97.77\%&  96.84\%&  \textbf{81.94\%}&  59.38\%& \textbf{67.90\%}&\textbf{85.07\%}\\
     \midrule
        \textbf{ST2V}         &192& 75.02\% & 87.93\% & 95.96\% & 94.71\% & \textbf{80.56}\% & 48.08\% & 57.85\% &77.16\%\\
        \textbf{Ours}         &192&  \textbf{91.32\%}&  \textbf{92.93\%}&  \textbf{97.40\%}&  \textbf{95.55\%}&  {77.78\%}&  \textbf{55.49\%}& \textbf{66.93\%} &\textbf{82.49\%}\\
        
        \bottomrule
        \end{tabular}
    }
    \caption{Evaluation metrics:
Comparison of performance metrics for various video generation methods as benchmarked by VBench. 
Bold values represent the best performance within each group.
}
    \label{tab:metrics}
    
\end{table}
Table~\ref{tab:metrics} reveals that our method, when compared to our base model VideoCrafter 2~\cite{chen2024videocrafter2}, experiences a slight decrease in most metrics except for the metric of dynamic.
In the generation of 64-frame videos, the performance of our method shows mixed results compared to other methods, with both advantages and disadvantages noted. 
However, our average metric scores are higher than those of both FreeNoise and OpenSora V1.1.
In the generation of longer 192-frame videos, our method outperforms StreamingT2V, the only other method capable of producing videos of this length, across the majority of evaluated metrics.

\subsection{Ablation}
\begin{figure}[t]
    \centering
    \includegraphics[page=1, width=1\textwidth]{image_placeholder/ablation.pdf}
    \vspace{-1em}
    \caption{
    Visualization of ablation studies on temporal module communication and context effects in video generation. 
    Top panel: Ablation of communication between the ResLayer module and the Attention module, showcasing two adjacent frames from the video sequence generated on different GPUs. 
    Bottom panel: Effects of ablating different contexts within the Attention module, displaying frames from videos generated post-ablation. 
    }
    \label{fig:ablation}
    \vspace{-1em}
\end{figure}

As mentioned in Section~\ref{section:module_parallel}, three types of temporal modules (\codeform{Conv()}, \codeform{GroupNorm()}, and \codeform{Attention()}) are adapted to synchronize context in \pname. 
To demonstrate the effectiveness of context synchronization by these modules, we conducted ablation experiments and visualized in Figure~\ref{fig:ablation} the impact of removing certain parts of the context synchronization on the quality of the generated videos. 
We performed ablation on the communication between the temporal \codeform{Attention()} module and the temporal \codeform{ResNet()} module in the video diffusion model, where the ResNet module includes temporal \codeform{Conv()}  and temporal \codeform{GroupNorm()} as submodules. 
Subsequently, we conducted ablations on the global  context $c_\text{global}$and the local context $c^i_\text{pre}, c^i_\text{post}$within the \codeform{Attention()} module.

\noindent\textbf{Removing local context.} From the top panel of Figure~\ref{fig:ablation}, it can be observed that the absence of synchronized information from the \codeform{ResNet()} leads to discrepancies in detail between the last frame on \codeform{device(1)} (frame 23) and the first frame on \codeform{device(2)} (frame 24), which are highlighted in red. 
These discrepancies, such as differences in the color of the clothes of the person behind the robot and the shape of the parts in the robot's hands on the table, do not appear in the original inference. 
When context of the \codeform{Attention()} module is absent, frame 23 and frame 24 become markedly different images, illustrating a significant discontinuity between video segments generated on adjacent devices. 
These observations suggest that synchronization in both \codeform{ResNet()} and \codeform{Attention()} modules is crucial for preserving visual coherence and continuity across video frames generated on different devices.

\noindent\textbf{Removing global context.} The bottom panel of Figure~\ref{fig:ablation} demonstrates that when synchronization of the global context is absent, content consistency within the video is difficult to maintain. 
For example, in frames 12 and 16, the horizon remains high, but in frames beyond 20, there is a noticeable rise in the horizon. 
Furthermore, when the local context synchronization is removed, although the content across different device clips remains consistent, the lack of shared context in the transition areas leads to anomalies. 
For instance, the content of snow in frame 22 abruptly transitions to a dog, highlighted in red. 
These examples highlight the importance of global and local context synchronization for video generation.

\section{Conclusion}

We presented {\methodname}, a distributed inference pipeline that leverages multiple GPUs for long-form video generation. 
We present two mechanisms, {\pname} and {\emph{\attnname}}, to addressed key challenges associated with distributed video generation. 
{\pname} reduces communication overhead by optimizing the exchange of context information, while {\emph{\attnname}} modified self-attention to ensure coherence across devices. 
Together, these innovations enable the rapid generation of videos up to 2,300 frames long, vastly improving generation speeds compared to existing methods. 
This approach not only extends the practical utility of diffusion models for video production but also sets a new benchmark for efficiency in long-form video generation.

\section{Limitation}
To fully harness the potential of our method, it relies on the availability of multiple GPUs.
Additionally, our approach does not effectively handle video generation involving scene transitions.

{
    \small
    \bibliographystyle{plain}
    \bibliography{main}
}

\appendix
\newpage
\section{Appendix}

\subsection{Communication overhead}

Table~\ref{tab:overhead} demonstrates the additional time overhead caused by communication between different temporal modules. 
The experiments were conducted on multiple Nvidia A5000 GPUs, with two settings: a dual-GPU configuration and an eight-GPU configuration.

\begin{table}[h]
    \centering
    \small
        \begin{tabular}{l|cc}
        \toprule
             \textbf{Sync}&  \multicolumn{2}{c}{\textbf{Inference Time (s)}}\\
             & \textbf{2$\times$GPU} &  \textbf{8 $\times$ GPU} \\
        \midrule
             Plain&  145.4 &  149.5\\
        \midrule
             + \texttt{Conv()} &  152.9 ($5.1\%\uparrow$)&  157.1 ($5.1\%\uparrow$)\\
             + \texttt{GroupNorm()} &  158.3 ($8.9\%\uparrow$)&  160.1 ($7.1\%\uparrow$)\\
             + \texttt{Attention()} &  170.7 ($17.4\%\uparrow$)&  180.2 ($20.5\%\uparrow$)\\
             Full Sync&  182.3 ($25.3\%\uparrow$)&  192.3 ($28.6\%\uparrow$)\\
         \bottomrule
        \end{tabular}
    \caption{Effect of Synchronization on Inference Time}
    \label{tab:overhead}
\end{table}

\subsection{Communication Algorithm}
\label{ap:algorithm}
\begin{algorithm}
\caption{Distributed Temporal Module Communication}
\label{alg:distributed_temporal}
\begin{algorithmic}[1]
    \REQUIRE 
    $i$ (the ID of the device), 
    $v_{\text{in}}^{i}$ (the input latent segment)
    \ENSURE Seamless and efficient distribution of frames for video processing.

    \STATE Prepare the global context $c^i_\text{global}$ using $v_{\text{in}}^{i}$ 
    \STATE \texttt{dist.all\_gather($c^i_\text{global}$)}

    \IF{$i$ mod 2 == 1}
        \STATE $c^i_\text{pre}$ = \texttt{dist.recv(i+1)}
        \STATE Prepare the local context for \texttt{device(i+1)} using $v_{\text{in}}^{i}$ 
        \STATE \texttt{dist.send($c^{i+1}_\text{post}$)}
        
        \STATE $c^i_\text{post}$ = \texttt{dist.recv(i-1)}
        \STATE Prepare the local context for \texttt{device(i-1)} using $v_{\text{in}}^{i}$ 
        \STATE \texttt{dist.send($c^{i-1}_\text{pre}$)}
        
    \ELSE
        \STATE $c^i_\text{post}$ = \texttt{dist.recv(i-1)}
        \STATE Prepare the local context for \texttt{device(i-1)} using $v_{\text{in}}^{i}$ 
        \STATE \texttt{dist.send($c^{i-1}_\text{pre}$)}
        
        \STATE $c^i_\text{pre}$ = \texttt{dist.recv(i+1)}
        \STATE Prepare the local context for \texttt{device(i+1)} using $v_{\text{in}}^{i}$ 
        \STATE \texttt{dist.send($c^{i+1}_\text{post}$)}
    \ENDIF
\end{algorithmic}
\end{algorithm}

\section{Gallery}
More videos are available in the supplementary materials and at the following link.
\href{https://ubiquitous-lobe-604.notion.site/Videos-Gallery-57dd5b23506f483d9f28dc27547b877a}{Link}


\end{document}